\documentclass[10pt,twocolumn,letterpaper]{article}

\usepackage{cvpr}
\usepackage{times}
\usepackage{epsfig}
\usepackage{graphicx}
\usepackage{amsmath}
\usepackage{amssymb}

\usepackage{multirow}
\usepackage{subfigure}

\usepackage[breaklinks=true,bookmarks=false]{hyperref}

\cvprfinalcopy 

\def\cvprPaperID{****} 

\begin{document}

\title{Knowledge-Embedded Routing Network for Scene Graph Generation}

\author{Tianshui Chen\\
Sun Yat-Sen University\\
DarkMatter AI Research\\
{\tt\small tianshuichen@gmail.com}
\and
Weihao Yu\\
Sun Yat-Sen University\\
{\tt\small weihaoyu6@gmail.com}\\
\and
Riquan Chen\\
Sun Yat-Sen University\\
{\tt\small sysucrq@gmail.com}
\and
Liang Lin\thanks{Tianshui Chen and Weihao Yu share first-authorship. Corresponding author is Liang Lin.}\\
Sun Yat-Sen University\\
DarkMatter AI Research\\
{\tt\small linliang@ieee.org}
}

\maketitle

\begin{abstract}
To understand a scene in depth not only involves locating/recognizing individual objects, but also requires to infer the relationships and interactions among them. However, since the distribution of real-world relationships is seriously unbalanced, existing methods perform quite poorly for the less frequent relationships. In this work, we find that the statistical correlations between object pairs and their relationships can effectively regularize semantic space and make prediction less ambiguous, and thus well address the unbalanced distribution issue. To achieve this, we incorporate these statistical correlations into deep neural networks to facilitate scene graph generation by developing a Knowledge-Embedded Routing Network. More specifically, we show that the statistical correlations between objects appearing in images and their relationships, can be explicitly represented by a structured knowledge graph, and a routing mechanism is learned to propagate messages through the graph to explore their interactions. Extensive experiments on the large-scale Visual Genome dataset demonstrate the superiority of the proposed method over current state-of-the-art competitors.
\end{abstract}

\section{Introduction}
Scene graph \cite{johnson2015image} is a structured representation of image content that not only encodes semantic and spatial information of individual objects in the scene but also represents the relationship between each pair of objects. In recent years, inferring such graph has drawn increasing attentions \cite{xu2017scene,dai2017detecting} as it provides a deeper understanding for the image and thus facilitates various vision tasks ranging from fundamental recognition and detection \cite{marino2017more,fang2017object} to high-level tasks \cite{zitnick2013learning,yatskar2016situation}. 

\begin{figure}[!t]
\centering
\subfigure[]{
\includegraphics[width=0.47\linewidth]{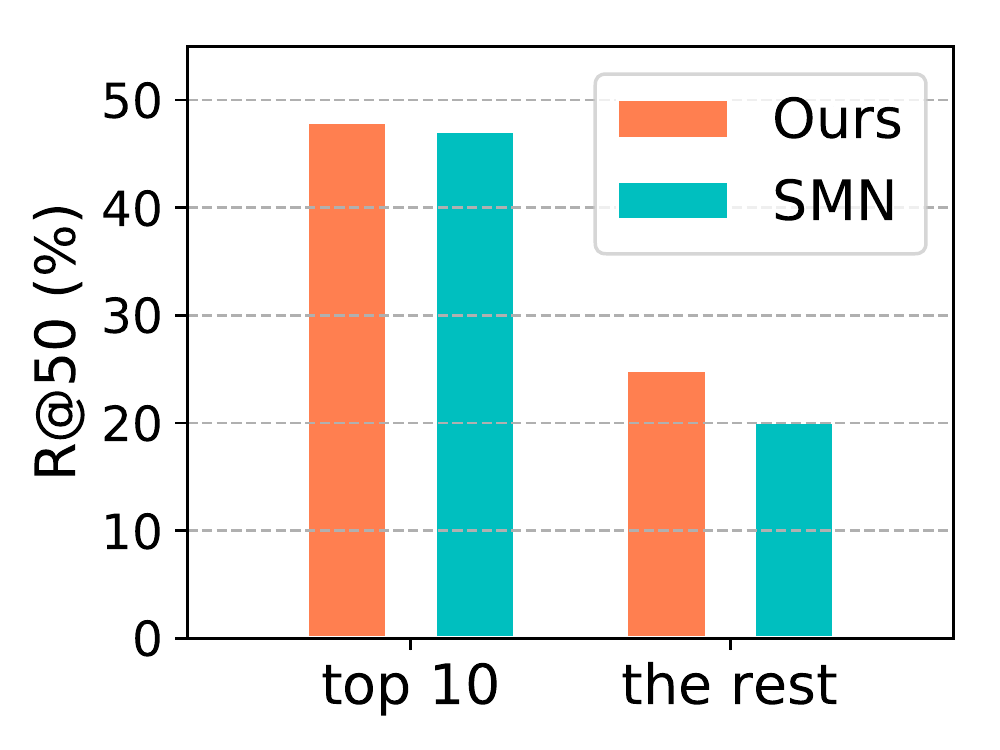}
\label{fig:spilt-evaluation-r50}}
\subfigure[]{
\includegraphics[width=0.47\linewidth]{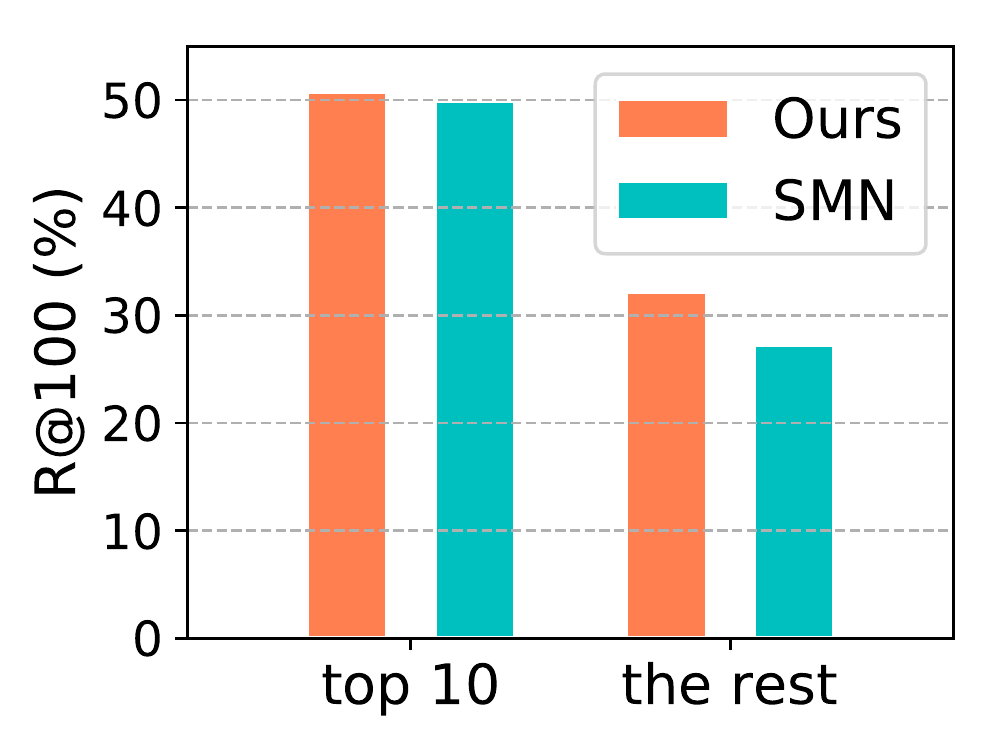}
\label{fig:spilt-evaluation-r100}}
\caption{(a) Recall@50 and (b) Recall@100 of our proposed method and the SMN \cite{zellers2017neural} on the scene graph classification task on the Visual Genome dataset \cite{krishna2017visual}. Both models are trained on the whole training set and evaluated on the two subsets, respectively. Note that SMN is the previous best-performing method.}
\label{fig:spilt-evaluation}
\end{figure}

Existing methods for scene graph generation rely on the target object regions \cite{lu2016visual, dai2017detecting} or further introduce contextual cues \cite{xu2017scene,zellers2017neural} to aid recognition. Generally, these methods require large amounts of annotated samples for model optimization. However, the distribution of real-world relationships is seriously uneven, leading to relatively poor performance for the relationships with limited training samples. Take the Visual Genome dataset \cite{krishna2017visual} as an example, we evaluate the performance on samples of top 10 most frequent relationships (namely ``top 10'' subset) and that on samples of the rest less frequent relationships (namely ``the rest'' subset), respectively. As shown in Figure \ref{fig:spilt-evaluation}, current best-performing method (i.e, SMN \cite{zellers2017neural}) can achieve competitive performance if it has sufficient training samples, but its performance suffers from a severe drop otherwise.

Objects in visual scene commonly have strongly structured regularities \cite{zellers2017neural}. For example, people tend to wear clothes, while cars are inclined to have wheels. The statistical analysis \cite{zellers2017neural} on the Visual Genome dataset \cite{krishna2017visual} revealed that a baseline method, which directly predicts the most frequent relationship of object pairs with given labels, outperforms most existing state-of-the-art methods \cite{newell2017pixels,xu2017scene}. Therefore, modeling these statistical correlations between object pairs and relationships can effectively regularize the semantic prediction space, and thus address the uneven distribution issue. On the other hand, the interplay of relationships and objects in the scene also plays a significant role in scene graph generation \cite{xu2017scene}. 

We show that the statistical correlations between object pairs and their relationships can be explicitly represented by a structured knowledge graph, and the interplay between these two factors can be captured by propagating node messages through the graph. Similarly, contextual cues can also be represented and explored by another graph with proper message propagation. In this work, we introduce a novel Knowledge-Embedded Routing Network (KERN), which captures the interplay of target objects and their relationships under the explicit guidance of prior statistical knowledge and automatically mines contextual cues to facilitate scene graph generation. Although previous studies \cite{dai2017detecting,zellers2017neural} have also taken notice of the statistical knowledge, they merely implicitly mine this information by iterative message propagation between relationships and objects \cite{xu2017scene} or by encoding the global context of objects and relationships \cite{zellers2017neural}. Instead, our model formally represents this statistical knowledge in the form of a structured graph and incorporates the graph into deep propagation network as extra guidance. In this way, it can effectively regularize the distribution of possible relationships of object pairs and thus make prediction less ambiguous. As shown in Figure \ref{fig:spilt-evaluation}, compared with current best-performing method (i.e., SMN \cite{zellers2017neural}), our model achieves slight improvement for the relationships with sufficient samples, and the improvement is much more evident for the relationships with limited samples.

Our model builds on the Faster RCNN detector \cite{ren2015faster} to generate a set of object regions. Then, a graph that correlates these regions according to the statistical object co-occurrences is first built, and a propagation network is employed to propagate node messages through the graph to learn contextualized feature representation to predict the class label regarding each region. For each object pair with predicted labels, we build a graph, in which nodes represent the objects and relationships, and edges represent the statistical co-occurrence probabilities between the given object pair and all relationships. Further, we adopt another propagation network to explore the interplay between the relationships and corresponding objects to predict their relationship. This process is performed for all object pairs, and the whole scene graph is generated.

On the other hand, existing works utilize the recall@$K$ (short as R@$K$) \cite{lu2016visual} as the evaluation metric. However, this metric is easily dominated by the performance of the relationships with a large proportion of samples. As the distribution of different relationships is severely uneven, if one method performs well on several most frequent relationships, it can achieve a high R@$K$ score. Thus, it can not well measure the performance of all relationships. To address this issue, we further propose a mean recall@$K$ (short as mR@$K$) as a complimentary evaluation metric. It first computes the R@$K$ for samples of each relationship and then averages over all relationships to obtain mR@$K$. Compared with R@$K$, mR@$K$ can give a more comprehensive performance evaluation for all relationships.

To the best of our knowledge, this work is the first to explicitly unify the statistical knowledge with the deep architecture to facilitate scene graph generation. Compared with existing methods, our model incorporates this knowledge to regularize the semantic space of relationship prediction and thus improves the performance of scene graph generation. We conduct experiments on the most widely used and challenging Visual Genome dataset \cite{krishna2017visual}, and demonstrate our model can achieve best R@$K$ performance than existing leading competitors. Notably, by explicitly regularizing the semantic space of relationship prediction, our model can well address the issue of uneven distribution of real-world relationships and achieves much more obvious improvement on the mR@$K$ metric. For example, our model improves the mR@50 and mR@100 from 15.4\% and 20.6\% to 19.8\% and 26.2\% on the scene graph classification task, with relative improvements of 28.6\% and 27.2\%, respectively. 

\section{Related Work}

\subsection{Visual relationship detection}
Visual relationship detection involves detecting semantic objects that occur in the images and inferring the relationship between each object pair (i.e., a subject and an object). Over the past decade, a series of works were dedicated to recognizing spatial relationships \cite{galleguillos2008object,gould2008multi,choi2013understanding} like ``above'', ``below'', ``inside'', and ``around'', and to exploring using these relationships to improve various vision tasks such as object recognition \cite{galleguillos2008object}, detection \cite{fang2017object}, and segmentation \cite{gould2008multi}. Some other works also attempted to learn human-object interactions \cite{yao2010grouplet,chao2017learning}, in which the subject was a person. 

Latterly, lots of attentions \cite{lu2016visual,xu2017scene,dai2017detecting,li2017scene,newell2017pixels,zellers2017neural,newell2017pixels} were drawn to the visual relationship detection task under a more general and practical setting, where the subject and object can be any objects in the scene and their relationships cover a wide range of relationship types including spatial (e.g., above, below), actions (e.g., ride, wear), affiliations (e.g., part of), etc. As a pioneer work, Lu et al. \cite{lu2016visual} trained visual models of subject, relationship, and object individually to tackle the problem of the long-tail distribution of relationship triplets and leveraged language prior from semantic word embedding to further improve the predicted performance. Xu et al. \cite{xu2017scene} introduced an end-to-end model that learned to iteratively refine relationship and object prediction via message passing based on the RNNs \cite{mikolov2010recurrent}. Li et al. \cite{li2017scene} formulated a multi-task framework to explore semantic associations over three tasks of object detection, scene graph generation, and image caption generation, and found that jointly learning the three tasks could bring about mutual improvements. More recently, Dai et al. \cite{dai2017detecting} designed a deep relational network that exploited both spatial configuration and statistical dependency to resolve the ambiguities during relationship recognition. Zeller et al. \cite{zellers2017neural} presented an analysis of statistical co-occurrences between relationships and object pairs on the Visual Genome dataset \cite{krishna2017visual} and came to a conclusion that these statistical co-occurrences provided strong regularization for relationship prediction. They encoded the global context of objects and relationships by LSTM sequential architectures \cite{hochreiter1997long} to facilitate scene graph parsing. 

The works \cite{dai2017detecting,zellers2017neural} also took notice of the statistical co-occurrences between object pair and their relationship, but they devised deep models to implicitly mine this information via message passing. Different from these works, our model formally represents this information and explicitly incorporates them into graph propagation network to help scene graph generation. 

\subsection{Knowledge representation}
It has been extensively studied to incorporate prior knowledge to aid numerous vision tasks \cite{marino2017more,fang2017object,lee2017multi,deng2014large,chen2018neural,lin2017knowledge}. For example, Marino et al. \cite{marino2017more} constructed a knowledge graph based on the WordNet \cite{miller1995wordnet} and the Visual Genome dataset \cite{krishna2017visual}, and learned the representation of this graph to enhance image feature representation to promote multi-label recognition. Lee et al. \cite{lee2017multi} further extended this method to multi-label zero-shot learning. Some works also utilized the knowledge graph as extra constraints for model training. Fang et al. \cite{fang2017object} incorporated semantic consistency into object detection systems with the constraint that more semantically consistent concepts were more likely to occur in an image. Deng et al. \cite{deng2014large} introduced semantic relations including mutual exclusion, overlap, and subsumption, as constraints in the loss function to train the classifiers. These methods learned graph representation for feature enhancement or use graph as extra constraints on the loss functions. Differently, our model introduces the graph that correlates target object pair and their possible relationships to explicitly regularize the semantic space of relationship prediction, and thus addresses the uneven distribution issue.

\section{Proposed Model}
A scene graph is a structured representation of content in an image. It consists of the class labels and locations of individual objects and the relationship between each object pair, which can be defined as a 3-tuple set $\mathcal{G}=\{B, O, R\}$:
\begin{itemize}
  \item $B=\{b_1, b_2, \dots, b_n\}$ is the region candidate set, with element $b_i \in \mathbb{R}^4$ denoting the bounding box of the $i$-th region.  
  \item $O=\{o_1, o_2, \dots, o_n\}$ is the object set, with element $o_i \in \mathbb{N}$ denoting the corresponding class label regarding region $b_i$. 
  \item $R=\{r_{1\rightarrow 2}, r_{1\rightarrow 3}, \dots, r_{n\rightarrow n-1}\}$ is the corresponding relationship triplet set, where $r_{i \rightarrow j}$ is a triplet of a subject $(b_i, o_i) \in B \times O$, an object $(b_j, o_j) \in B \times O$, and a relationship label $x_{i \rightarrow j} \in \mathcal{R}$.
\end{itemize}
$\mathcal{R}$ is the set of all relationships including \emph{no-relationship} that indicates no relationship between the given object pair.

\begin{figure}[!t]
\centering
\subfigure[]{
\includegraphics[width=0.57\linewidth]{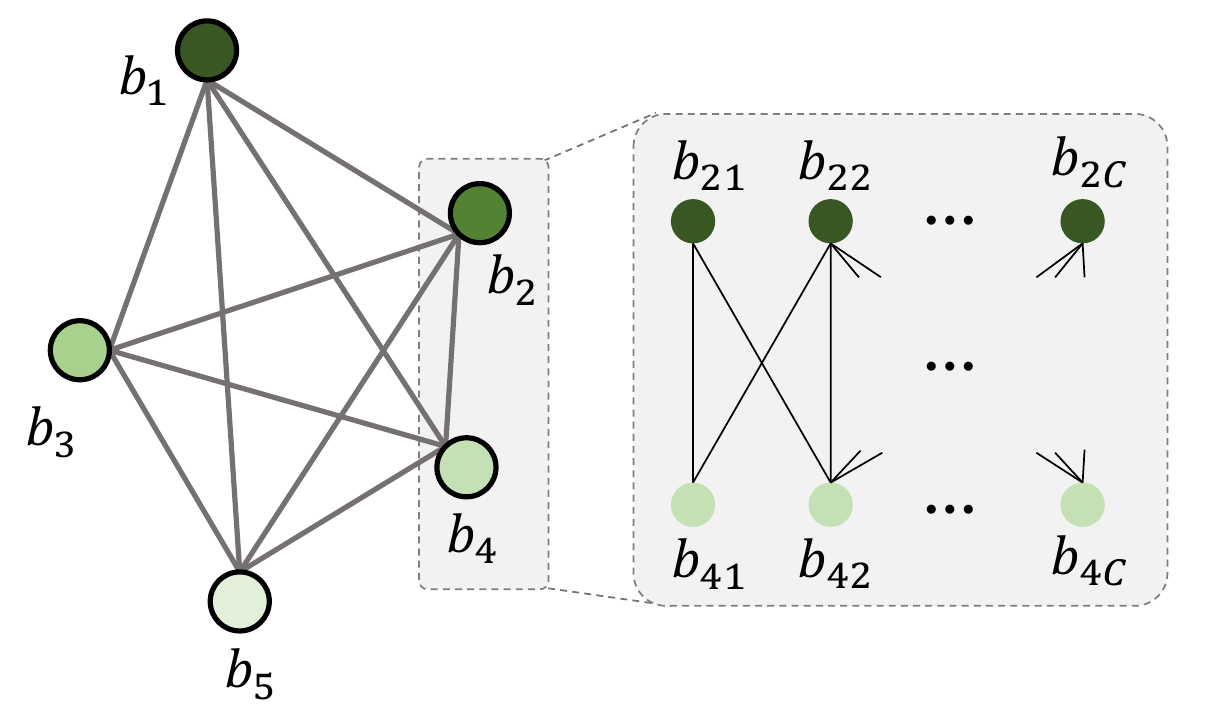}
\label{fig:graph1}}
\subfigure[]{
\includegraphics[width=0.37\linewidth]{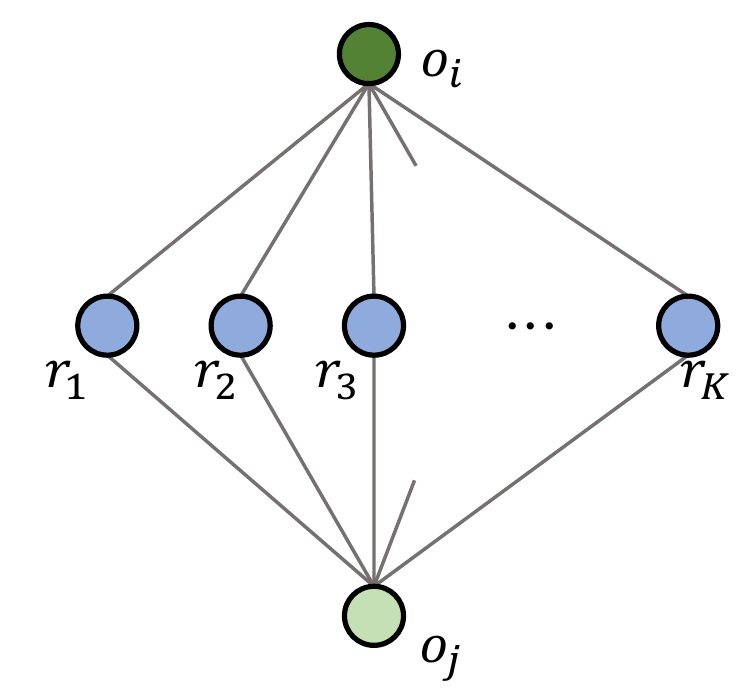}
\label{fig:graph2}}
\caption{(a) A graph correlating the detected regions appearing in an image; (b) A graph correlating given object pair $o_i$ and $o_j$ with all the relationships.}
\end{figure}

\begin{figure*}[!t]
   \centering
   \includegraphics[width=1.0\linewidth]{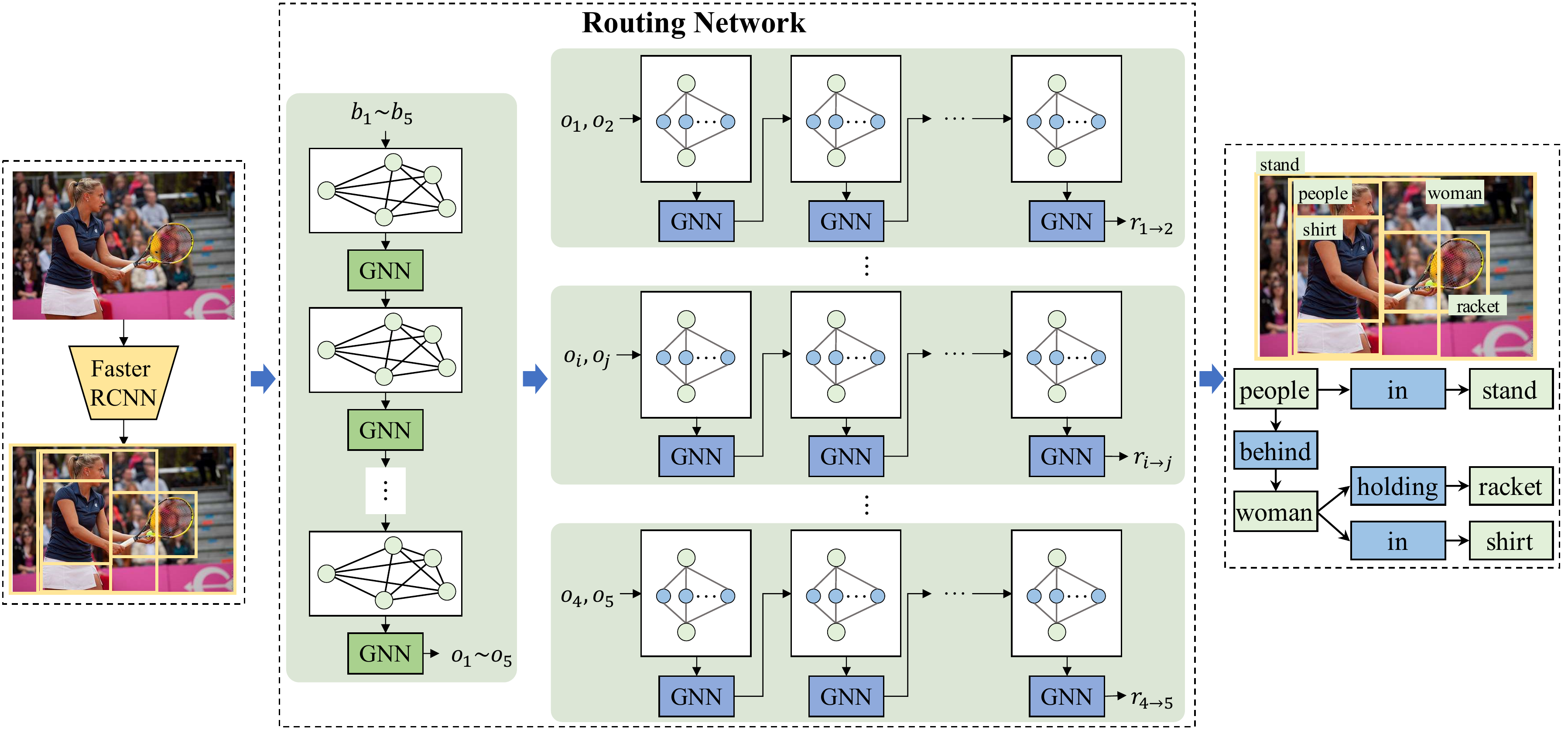} 
   \caption{An overall pipeline of the knowledge-embedded routing network. Given an image, we first adopt the Faster RCNN to detect a set of regions. Then, a graph is built to correlate the regions, and a graph neural network is employed to learn contextualized representation to predict the class label for each region. For each object pair with predicted labels, we build another graph to correlate the given object pair with all the possible relationships and employ a graph neural network to infer their relationship. The process is repeated for all object pairs and the scene graph is generated.}
   \label{fig:framework}
\end{figure*}

Given an image $I$, we decompose the probability distribution of the scene graph $p(\mathcal{G}|I)$ into three components similar to \cite{zellers2017neural}:

\begin{equation}
  p(\mathcal{G}|I)=p(B|I)p(O|B,I)p(R|O,B,I).
\end{equation}
In this equation, the bounding box component $p(B|I)$ generates a set of candidate regions that cover most of the key objects directly from the input image. Similar to previous scene graph works \cite{dai2017detecting,zellers2017neural}, this component is implemented by the widely used Faster RCNN detector \cite{ren2015faster}. The object component $p(O|B,I)$ then predicts the class label regarding each detected region. Here, we construct a graph that correlates the detected regions based on the statistical object co-occurrence information (see Figure \ref{fig:graph1}). Then, our model adopts a graph neural network \cite{scarselli2009graph,li2016gated} to propagate messages through the graph to learn contextualized representation for each region and achieves better label prediction under the constraint of statistical information of object co-occurrences. Conditioned on the predicted labels, the relationship component $p(R|O,B,I)$ infers the relationship of each object pair and finally generates the whole scene graph. For each object pair with predicted labels, we construct a graph, in which nodes refer to the objects and relationships, and edges represent the statistical co-occurrences between the corresponding object pair and all the relationships (see Figure \ref{fig:graph2}). Similarly, another graph neural network is learned to explore the interplay between relationships and objects, and finally, the features from all nodes are aggregated to predict the relationship. Our model performs this process for all object pairs and generates the whole scene graph. Figure \ref{fig:framework} illustrates an overall pipeline of the proposed model.

\subsection{Bounding box localization}
\label{sec:bbl}
Given an image, the model first obtains a set of candidate regions. In this work, we utilize the Faster RCNN \cite{ren2015faster} to automatically generate the region set $B=\{b_1, b_2, \dots, b_n\}$ directly from input image $I$. For each region, besides a bounding box $b_i \in \mathbb{R}^4$ denoting its position, our model also extracts a feature vector $\mathbf{f}_i$ using the ROI pooling layer \cite{girshick2015fast}. These feature vectors are then fed into the propagation networks for subsequent inference.

\subsection{Knowledge-embedded routing network}
\label{sec:spn}

\noindent\textbf{Object. }Statistical information of object co-occurrence is a crucial cue to correlate objects in an image and regularizes object label prediction. In this work, we build a graph to associate the regions detected in the image according to these statistical correlations and employ a graph neural network to propagate messages through the graph that can learn contextualized representation to predict the class label regarding each region.

To this end, we first count the statistical co-occurrence probabilities of objects from different categories on the training set of the target dataset (e.g., Visual Genome \cite{krishna2017visual}). More specifically, for two categories of $c$ and $c'$, we count the probability $m_{cc'}$ of the existence of object belonging to category $c$ in the presence of object belonging to category $c'$. We count these co-occurrence probabilities for all category pair and obtain a matrix $M_c \in \mathbb{R}^{C \times C}$, where $C$ is the number of object categories. We then correlate the regions from $B$ based on the matrix $M_c$. Given two regions of $b_i$ and $b_j$, we duplicate $b_i$ $C$ times to obtain $C$ nodes $\{b_{i1}, b_{i2}, \dots, b_{iC}\}$, with node $b_{ic}$ denoting the correlation of region $b_i$ with category $c$.  The same process is performed for $b_j$. Intuitively, $m_{{c}{c'}}$ can be used to correlate node $b_{j{c'}}$ to $b_{i{c}}$, and thus $M_c$ can be used to correlate nodes of region $b_i$ and nodes of $b_j$. In this way, we can correlate all regions and construct the graph.

Inspired by the Graph Gated Neural Networks \cite{li2016gated,chen2018knowledge,wang2018deep}, we adopt a gated recurrent update mechanism to iterative propagate node messages through the graph. Specifically, at timestep $t$, each node $b_{ic}$ has a hidden state $\mathbf{h}_{ic}^t$. As each node corresponds to a specific region, we use the feature vector of this region to initialize the hidden state at $t=0$, which can be expressed as
\begin{equation}
  \mathbf{h}_{ic}^0=\varphi_o(\mathbf{f}_i),
  \label{eq:object-initialization}
\end{equation}
where $\phi_o$ is a transformation that maps $\mathbf{f}_i$ to a feature vector of low dimension, and it is implemented by a fully connected layer. At each timestep $t$, each node aggregates messages from its neighbors according to the graph structure, formulated as
\begin{equation}
  \mathbf{a}_{ic}^t=\left[\sum_{j=1,j\neq i}^{n}\sum_{c'=1}^{C}m_{c'c}\mathbf{h}_{jc'}^{t-1}, \sum_{j=1,j\neq i}^{n}\sum_{c'=1}^{C}m_{cc'}\mathbf{h}_{jc'}^{t-1}\right].
\end{equation}
Then, the model take $a_{ic}^t$ and its previous hidden state as input to update its hidden state by a gated mechanism similar to the Gated Recurrent Unit \cite{cho2014learning,li2016gated}
\begin{equation}
   \begin{split}
    \mathbf{z}_{ic}^t=&\sigma(\mathbf{W}_o^z{\mathbf{a}_{ic}^t}+\mathbf{U}_o^z{\mathbf{h}_{ic}^{t-1}}) \\
    \mathbf{r}_{ic}^t=&\sigma(\mathbf{W}_o^r{\mathbf{a}_{ic}^t}+\mathbf{U}_o^r{\mathbf{h}_{ic}^{t-1}}) \\
    \widetilde{\mathbf{h}_{ic}^t}=&\tanh\left(\mathbf{W}_o{\mathbf{a}_{ic}^t}+\mathbf{U}_o({\mathbf{r}_{ic}^t}\odot{\mathbf{h}_{ic}^{t-1}})\right)\\
    \mathbf{h}_{ic}^t=&(1-{\mathbf{z}_{ic}^t}) \odot{\mathbf{h}_{ic}^{t-1}}+{\mathbf{z}_{ic}^t}\odot{\widetilde{\mathbf{h}_{ic}^t}}\\
   \end{split}
   \label{eq:ggnn}
\end{equation}
In this way, each node can aggregate messages from its neighbors and meanwhile transfer its message to its neighbors, enabling interactions among all nodes in the graph. After $T_o$ steps, the node messages have been propagated through the graph and we obtain the final hidden state for each region $i$, i.e., $\{\mathbf{h}_{i1}^{T_o}, \mathbf{h}_{i2}^{T_o}, \dots, \mathbf{h}_{iC}^{T_o}\}$. We use an output network that takes the initial hidden state and final hidden state as input to compute the output feature for each node
\begin{equation}
  \mathbf{f}_{ic}^o=o_o(\mathbf{h}_{ic}^0, \mathbf{h}_{ic}^T),
\end{equation}
where $o_o(\cdot)$ is implemented by a fully connected layer. Finally, for each region, we aggregate all correlated output feature vectors to predict its class label
\begin{equation}
   \mathbf{o}_i=\phi_{o}(\mathbf{f}_{i1}^{o}, \mathbf{f}_{i2}^{o}, \dots, \mathbf{h}_{iC}^{o})\\
   \label{eq:object-cls}
\end{equation}
The predicted class label $o_i=\mathrm{argmax}(\mathbf{o}_i)$ are then used for relationship inference.

\noindent\textbf{Relationship. }Given the categories of object pair, the probability distribution of their relationships is highly skewed. For example, given a subject ``man'' and an object ``horse'', their relationship is likely to be ``riding''. Here, we represent the correlations of object pair and their relationships in the form of a structured graph and adopt another graph neural network to explore the interplay of these two factors to infer the relationship. 

To this, we also count the statistical co-occurrence probability on the training part of the target dataset to obtain these correlations. Concretely, we count the probabilities of all possible relationships given a subject of the category $c$ and an object of the category $c'$, which are denoted as $\{m_{cc'1}, m_{cc'2}, \dots, m_{cc'K}\}$. Here, $K$ is the relationship number. For a subject $o_i$ and an object $o_j$ taken from the object set $O$, we construct a graph with a subject node, an object node, and $K$ relationship nodes. We use $m_{{o_i}{o_j}{k}}$ to denote the correlations between $o_i$ and relationship node $k$ as well as between $o_j$ and relationship node $k$. In this way, a graph with statistic co-occurrences embedded is built.

Our model learns to explore the node interaction using the identical graph gated recurrent update mechanism \cite{li2016gated}. Similarly, each node $v \in V=\{o_i, o_j, 1, 2, \dots, K\}$ has a hidden state $\mathbf{h}_v^t$ at timestep $t$. At timestep $t=0$, we initialize the object nodes with the feature vectors of corresponding regions and the relationship nodes with the feature vector from the union region of the two objects together with their spatial information
\begin{equation}
\mathbf{h}^0_v=
\begin{cases}
\varphi_{o'}(\mathbf{f}_i) & \text{if $v$ is the object node $o_i$}\\
\varphi_r(\mathbf{f}_{ij}) & \text{if $v$ is a relationship node}
\end{cases},
\end{equation}
where $\varphi_{o'}$ and $\varphi_r$ are two transformations, and both are implemented by a fully-connected layer, respectively. $\mathbf{f}_{ij}$ is a feature vector that encodes the visual feature of the union region of $b_i$ and $b_j$ as well as the spatial information following \cite{zellers2017neural}. At each timestep $t$, the relationship nodes aggregate messages from the object nodes while object nodes aggregate messages from the relationship nodes
\begin{equation}
\mathbf{a}^t_v=
\begin{cases}
\sum_{k=1}^K m_{{o_i}{o_j}{k}}\mathbf{h}_{k}^{t-1}& \text{if $v$ is a object node}\\
 m_{{o_i}{o_j}{k}}(\mathbf{h}_{o_i}^{t-1}+\mathbf{h}_{o_j}^{t-1})& \text{if $v$ is the relationship node $k$}
\end{cases}.
\end{equation}
Then, the model incorporates these aggregated features with the previous hidden states to update the hidden state for each node using the gated mechanism as Eq. \ref{eq:ggnn}. The model repeats the iterations $T_r$ times and generates the final hidden state of each node, i.e., $\{\mathbf{h}_{o_i}^{T_r}, \mathbf{h}_{o_j}^{T_r}, \mathbf{h}_{1}^{T_r}, \dots, \mathbf{h}_{K}^{T_r},\}$. Similar to \cite{li2016gated}, our model use an output sub-network implemented by a fully-connected layer to compute node-level features and aggregates these features to infer the relationship
\begin{equation}
   \begin{split}
   \mathbf{f}_v^o&=o_r([\mathbf{h}_v^{T_r}, \mathbf{h}_v^0]) \\
   \mathbf{x}_{i \rightarrow j}&=\phi_{r}([\mathbf{f}_{o_i}^o, \mathbf{f}_{o_j}^o, \mathbf{f}_{1}^o, \dots, \mathbf{f}_{K}^o]).
   \end{split}
   \label{eq:relationship-cls}
\end{equation}
$\phi_{r}$ is the relationship classifier implemented by a fully connected layer.

\begin{table*}[!t]
\centering
\small
\begin{tabular}{c|c|cc|cc|cc|c}
\hline
\multirow{2}{*}{} & \centering \multirow{2}{*}{Method} & \multicolumn{2}{|c|}{SGGen}  & \multicolumn{2}{c|}{SGCls} & \multicolumn{2}{c|}{PredCls}  \\
 & & mR@50 & mR@100   & mR@50 & mR@100 &   mR@50 & mR@100 & Mean  \\
\hline
\hline
\multirow{6}{*}{Constraint} 
& \centering IMP \cite{xu2017scene} &  0.6   & 0.9 &  3.1 & 3.8 & 6.1 & 8.0 & 3.8 \\
& \centering IMP+ \cite{xu2017scene,zellers2017neural} &   3.8 & 4.8 & 5.8 & 6.0 & 9.8 & 10.5 & 6.8 \\
& \centering FREQ \cite{zellers2017neural} & 4.3 & 5.6 & 6.8 & 7.8 & 13.3 & 15.8 & 8.9\\
& \centering SMN \cite{zellers2017neural} & 5.3 & 6.1 & 7.1 & 7.6 & 13.3 & 14.4 & 9.0\\
& \centering \textbf{Ours} & \textbf{6.4} & \textbf{7.3} & \textbf{9.4} & \textbf{10.0} & \textbf{17.7} & \textbf{19.2} & \textbf{11.7}\\
\hline
\multirow{5}{*}{Unconstraint} 
& \centering AE \cite{newell2017pixels} & 1.6 & 2.5 & 6.0 & 7.8 & 15.1 & 19.5 & 8.8 \\
& \centering IMP+ \cite{xu2017scene,zellers2017neural} & 5.4 & 8.0 &  12.1 & 16.9 & 20.3 & 28.9 & 15.3\\
& \centering FREQ \cite{zellers2017neural} & 5.9 & 8.9 & 13.5 & 19.6 & 24.8 & 37.3 & 18.3 \\
& \centering SMN \cite{zellers2017neural} &9.3 & 12.9 & 15.4 & 20.6 & 27.5 & 37.9 & 20.6 \\
& \centering \textbf{Ours}  & \textbf{11.7} & \textbf{16.0} & \textbf{19.8} & \textbf{26.2} & \textbf{36.3} & \textbf{49.0} & \textbf{26.5}\\
\hline
\end{tabular}
\caption{Comparison of the mR@50 and mR@100 in \% with and without constraint on the three tasks of the VG dataset. We compute Mean mR by averaging mR@50 and mR@100 over the three tasks. As existing works do not present the mR@$K$ metric, we utilize the released models (IMP, FREQ, SMN, AE) or train the model using the released code (IMP+) to generate the results to compute the metric.}
\label{table:vg-mrk-sota}
\end{table*}

\section{Experiments}
\subsection{Experiment setting}

\noindent\textbf{Implementation details. }Similar to prior works \cite{xu2017scene,zellers2017neural} for scene graph generation, we adopt the Faster RCNN detector \cite{ren2015faster} to generate the candidate region set. The detector utilizes VGG16-ConvNet \cite{simonyan2015very} pretrained on ImageNet \cite{russakovsky2015imagenet} as its backbone network as in \cite{xu2017scene,zellers2017neural}. We follow \cite{zellers2017neural} to set the input image size as $592 \times 592$, and use anchor scales and aspect ratios similar to YOLO-9000 \cite{redmon2017yolo9000}. Then, we train the detector on the target dataset using the SGD algorithm with a batch size of 18, momentum of 0.9, and weight decay of 0.0001. The learning rate is initialized as 0.001 and is divided by 10 when the mAP of the validation set plateaus. After that, we freeze the weights of all the convolution layers and train the fully-connected layers as well as the stacked graph neural networks using the Adam algorithm with a batch size of 2, and momentums of 0.9 and 0.999. In this process, we initialize the learning rate as 0.00001 and divide it by 10 when the recall of the validation set plateaus.

\noindent\textbf{Datasets. }We evaluate the proposed method and existing state-of-the-art competitors on the Visual Genome (VG) \cite{krishna2017visual} benchmark. VG contains 108,077 images with average annotations of 38 objects and 22 relationships per image. It is a challenging and most widely used benchmark for scene graph generation. In the experiments, we follow previous works \cite{zellers2017neural,xu2017scene} to use the most frequent 150 object categories and 50 relationships and use the training/test split in \cite{xu2017scene} for evaluation.

\noindent\textbf{Tasks. }Scene graph generation aims to predict a set of \emph{subject-relationship-object} triplets. Following \cite{xu2017scene}, we evaluate the proposed model with three task setups as below: 

\begin{itemize}
  \item Predicate classification (PredCls) predicts the relationship label of given object pair from a set of objects with ground truth annotations of class labels and bounding boxes.  
  \item Scene graph classification (SGCls) predicts the class labels for the set of objects with ground truth bounding boxes and predicts the relationship label of each object pair. 
  \item Scene graph generation (SGGen) simultaneously detects objects appearing in the image and predicts the relationship label of each object pair.
\end{itemize}

\noindent\textbf{Evaluation metrics. }
All the methods are evaluated using the recall@$K$ (short as R@$K$) metric that measures the fraction of the ground truth relationship triplets that appear among the top $K$ most confident triplet predictions in an image. However, as shown in Figure \ref{fig:sg-distribution1}, the distribution of different relationships is seriously uneven, and this metric is easily dominated by the performance of the most frequent relationships. To evaluate the performance of each relationship more comprehensively, we further propose a new metric, i.e., mean recall@$K$ (short as mR@$K$). This metric computes the R@$K$ for the samples of each relationship, respectively, and then averages R@$K$ over all relationships to obtain mR@$K$.  

Some previous works \cite{xu2017scene} compute R@$K$ with the constraint that merely one relationship is obtained for a given object pair. Some other works \cite{newell2017pixels} omit this constraint so that multiple relationships can be obtained, leading to higher values. In this work, we report both the R@$K$ and mR@$K$ with and without constraint respectively for comprehensive comparisons.

\begin{table*}[!t]
\centering
\small
\begin{tabular}{c|c|cc|cc|cc|c}
\hline
\multirow{2}{*}{} & \centering \multirow{2}{*}{Methods} & \multicolumn{2}{|c|}{SGGen}  & \multicolumn{2}{c|}{SGCls} & \multicolumn{2}{c|}{PredCls}  \\
 & & \quad R@50 \quad  & R@100\quad & \quad R@50\quad  &  R@100\quad  &  \quad  R@50\quad  &  R@100 \quad & Mean  \\
\hline
\hline
\multirow{6}{*}{Constraint} 
& \centering VRD \cite{lu2016visual} &  0.3 & 0.5   &  11.8 & 14.1 &  27.9  & 35.0 & 14.9  \\
& \centering IMP \cite{xu2017scene} & 3.4 & 4.2   &    21.7 & 24.4 &   44.8 & 53.0   & 25.3\\
& \centering IMP+ \cite{xu2017scene,zellers2017neural}& 20.7 & 24.5  & 34.6 & 35.4 & 59.3 & 61.3 & 39.3 \\
& \centering FREQ \cite{zellers2017neural} & 23.5 & 27.6  & 32.4 & 34.0 & 59.9 & 64.1 & 40.3 \\
& \centering SMN \cite{zellers2017neural} & \textbf{27.2} & \textbf{30.3} & 35.8 & 36.5  & 65.2 & 67.1  & 43.7\\
& \centering \textbf{Ours} & 27.1 &29.8   & \textbf{36.7} & \textbf{37.4} &  \textbf{65.8} & \textbf{67.6}  & \textbf{44.1}  \\
\hline
\multirow{5}{*}{No constraint} 
& \centering AE \cite{newell2017pixels}   & 9.7 & 11.3 & 26.5 & 30.0 & 68.0  & 75.2 & 36.8 \\
& \centering IMP+ \cite{xu2017scene,zellers2017neural}&22.0  & 27.4   & 43.4 & 47.2 & 75.2 &  83.6 & 49.8\\
& \centering FREQ \cite{zellers2017neural}  &25.3 & 30.9 & 40.5 & 43.7 &  71.3 & 81.2 & 48.8 \\
& \centering SMN \cite{zellers2017neural} & 30.5 & 35.8 & 44.5 & 47.7  & 81.1 & 88.3& 54.7\\
& \centering \textbf{Ours}  & \textbf{30.9} & \textbf{35.8}  & \textbf{45.9} & \textbf{49.0}  &  \textbf{81.9} & \textbf{88.9} &\textbf{55.4}  \\
\hline
\end{tabular}
\caption{Comparison of the R@50 and R@100 in \% with and without constraint on the three tasks of the VG dataset. We compute Mean R by averaging R@50 and R@100 over the three tasks.}
\label{table:vg-sota}
\end{table*}


\subsection{Comparison with state-of-the-art methods}
VG \cite{krishna2017visual} is the largest and most widely used benchmark for evaluating the scene graph generation task. In this part, we compare our proposed method with the existing state-of-the-art methods, including Visual Relationship Detection (VRD) \cite{krishna2017visual}, Iterative Message Passing (IMP) \cite{xu2017scene} and its improved version by using a better detector (IMP+) \cite{xu2017scene,zellers2017neural}, Associative Embedding (AE) \cite{newell2017pixels}, FREQuency baseline (FREQ) \cite{zellers2017neural}, and Stacked Motif Networks (SMN) \cite{zellers2017neural}.

\begin{figure}[!t]
\centering
\subfigure[]{
\includegraphics[width=1\linewidth]{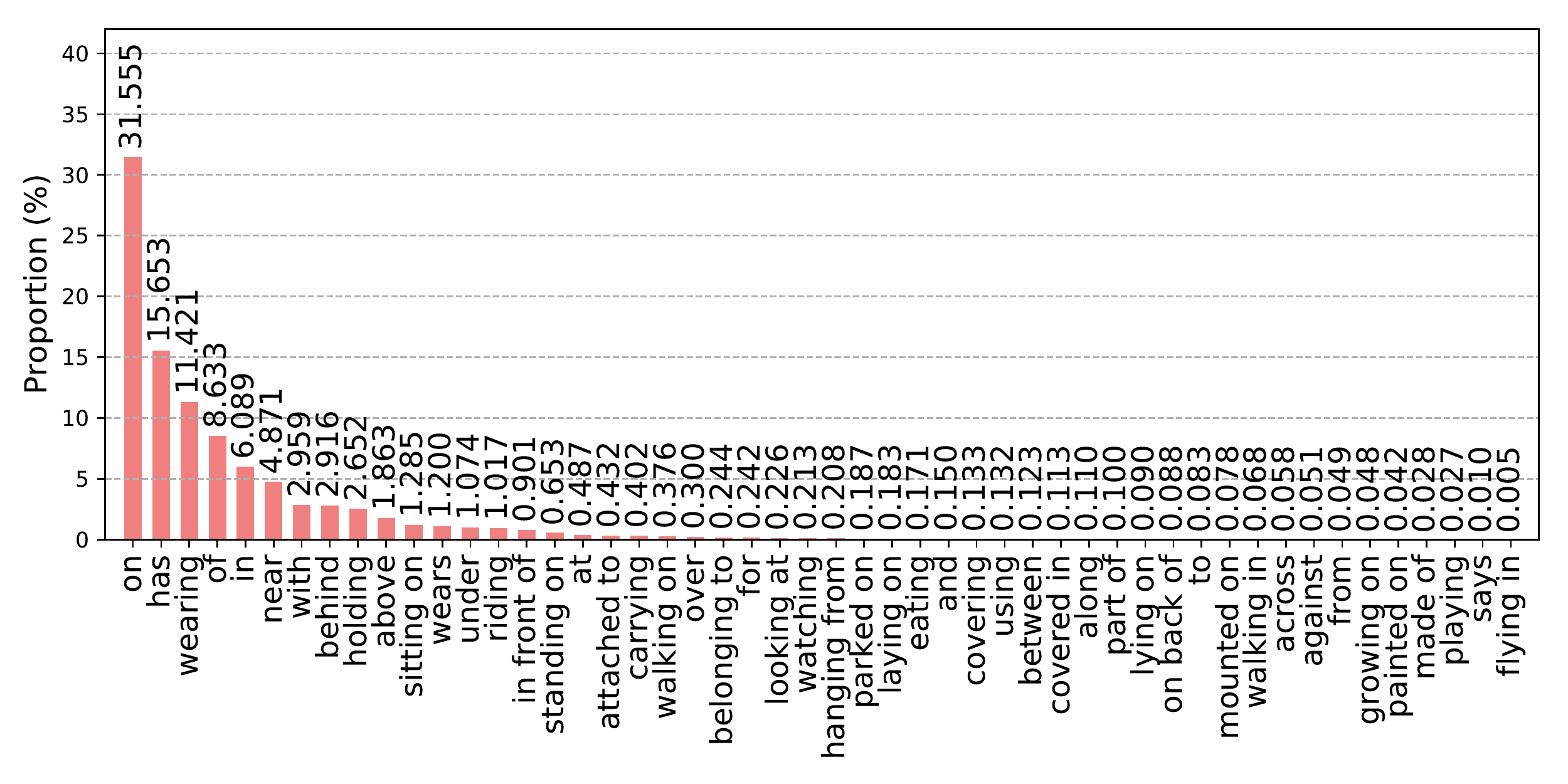}
\label{fig:sg-distribution1}}
\subfigure[]{
\includegraphics[width=1\linewidth]{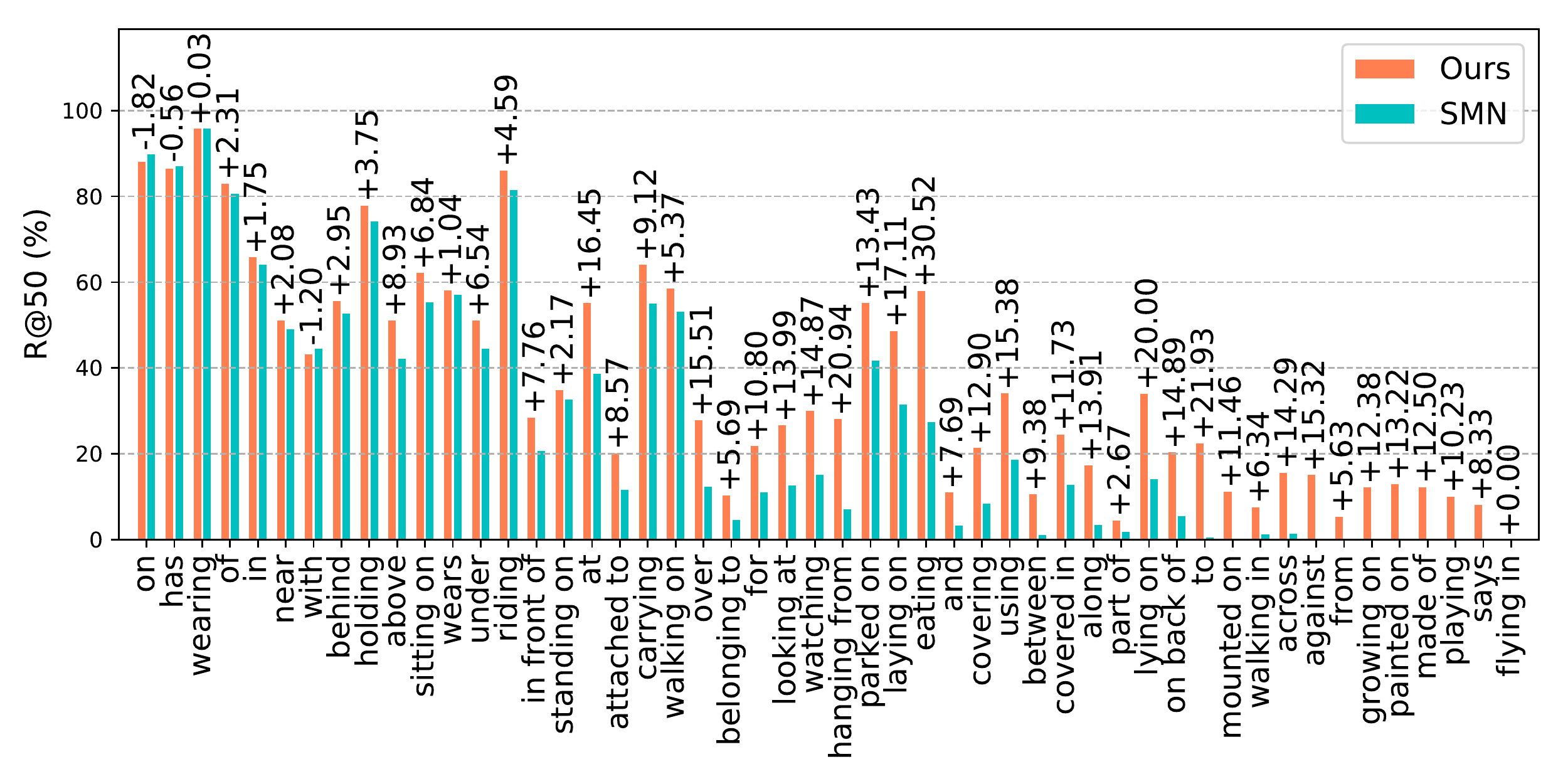}
\label{fig:sg-comparison}}
\caption{(a) The distribution of different relationships on the VG dataset. The training and test splits share similar distribution. (b) The R@50 without contraint of our method and the SMN on the predicate classification task on the VG dataset.}
\end{figure}

\begin{figure}[htp]
\centering
\subfigure[]{
\includegraphics[width=1\linewidth]{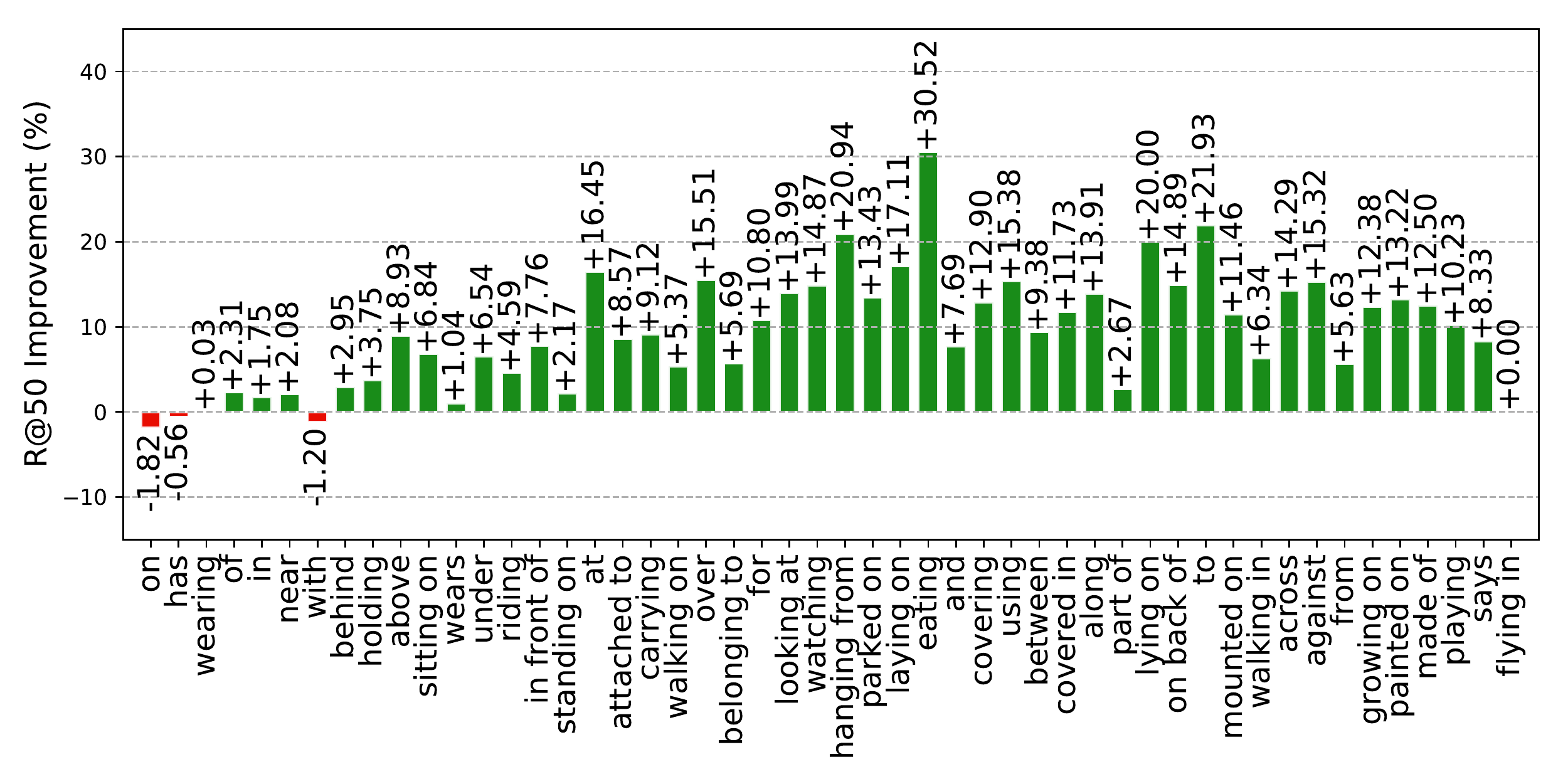}
\label{fig:sg-improvement}}
\subfigure[]{
\includegraphics[width=1\linewidth]{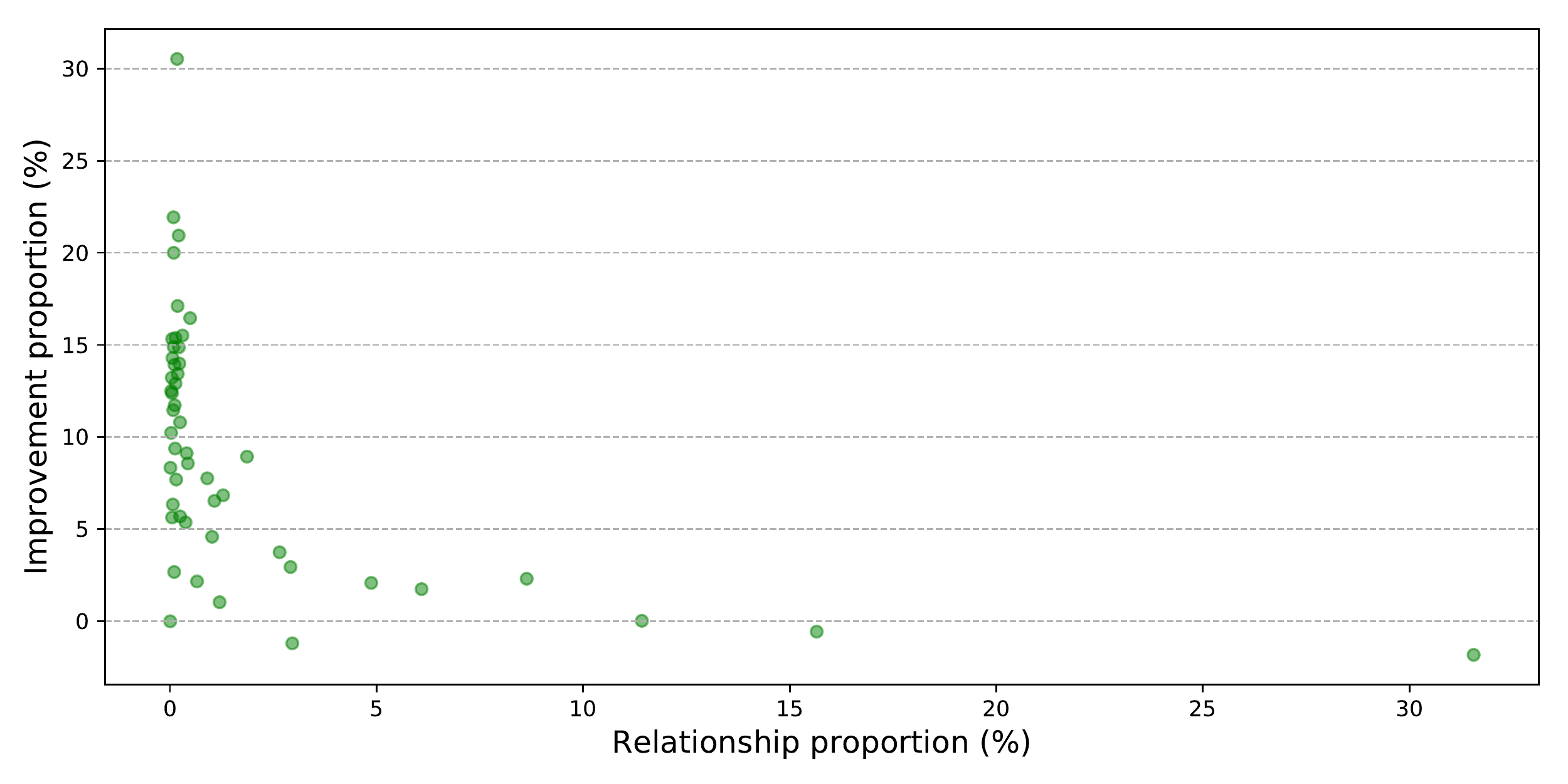}
\label{fig:sg-scatter}}
\caption{(a) The R@50 improvement of different relationships of our methods to the SMN and (b) the relation between the R@50 improvement and sample proportion on the predicate classification task on the VG dataset. The R@50 are computed without contraint.}
\end{figure}

\begin{table*}[htp]
\centering
\small
\begin{tabular}{c|cc|cc|cc|c}
\hline
  \centering \multirow{2}{*}{Methods} & \multicolumn{2}{|c|}{SGGen}  & \multicolumn{2}{c|}{SGCls} & \multicolumn{2}{c|}{PredCls}  \\
 & mR@50 & mR@100  & mR@50 & mR@100 & mR@50 & mR@100 & Mean  \\
\hline
\centering Ours w/o rk \& w/o ok & 5.1 & 5.8 & 6.1& 6.5 & 10.5 & 11.5 & 7.6\\
\centering Ours w/o rk &  5.2 & 5.9  & 6.5 & 6.9 & 11.1 & 12.0 & 7.9 \\
\centering Ours & \textbf{6.4} & \textbf{7.3} & \textbf{9.4} & \textbf{10.0} & \textbf{17.7} & \textbf{19.2} & \textbf{11.7}  \\
\hline
\hline 
& R@50 & R@100  & R@50 & R@100 & R@50 & R@100 & Mean  \\
\hline
\centering Ours w/o rk \& w/o ok & 25.2 & 27.9 & 33.9 & 34.8 & 58.7 & 61.0 & 40.3 \\
\centering Ours w/o rk & 25.5 & 28.0 & 34.3 & 35.2 & 59.2 & 61.5 & 40.6\\ 
\centering Ours & \textbf{27.1} & \textbf{29.8}  & \textbf{36.7} & \textbf{37.4}&  \textbf{65.8} & \textbf{67.6}& \textbf{44.1}    \\
\hline
\end{tabular}
\caption{Comparison of the mR@50, mR@100 (above) and the R@50, R@100 (below) with constraint in \% of our full model, our model without relationship correlation (w/o rc), and our model without relationship correlation and object correlation (w/o rc \& oc). We compute Mean mR by averaging mR@50 and mR@100 over the three tasks and mean R in the same way.}
\label{table:vg-mrk-ablation}
\end{table*}

We first present the mR@50 and mR@100 on three tasks on the VG dataset in Table \ref{table:vg-mrk-sota}. As shown, the  FREQ baseline method, which directly predicts the most frequent relationship of object pairs with given labels, performs better than most existing works. This comparison suggests that the statistical correlations between object pairs and their relationships play an equally or even more important role than other information like contextual cues \cite{xu2017scene}. SMN is the best-performing method among existing works, which implicitly captures these statistical correlations by encoding global context. It achieves the mean mR of 9.0\% and 20.6\% under the evaluation settings with and without constraint. By explicitly incorporating the statistical correlations, our method can make better use of them, leading to notable performance improvement. Specifically, it consistently outperforms existing methods on all three tasks under the two settings. For example, it obtains the mean mR of 11.7\% and 26.5\%, with a relative improvement of 30.0\% and 28.6\% compared with the previous best-performing method (i.e., SMN). Note that we use prior statistical correlations to aid scene graph generation. But these correlations are obtained merely based on the annotations of samples from the training set, and no additional supervision is introduced. Thus, the preceding comparisons are fair. 

For more comprehensive comparison with existing methods, we also present the R@50 and R@100 on the three tasks on the VG dataset in Table \ref{table:vg-sota}. Still, our method achieves best results on these metrics. Concretely, the mean R is 44.1\% and 55.4\% under the settings with and without constraint, with an improvement of 0.4\% and 0.7\% compared with SMN.

As shown in the above discussion and comparison, our method exhibits an improvement compared with existing state-of-the-art methods, both on the mR@$K$ and R@$K$ metrics. However, we find that the improvement on the mR@$K$ metric is much more obvious than that on the R@$K$ metric. Here, we give a deeper and more comprehensive analysis for this phenomenon. We first present the distribution of different relationships on the VG dataset in Figure \ref{fig:sg-distribution1}, and the corresponding distributions on the training and test splits are basically the same to this distribution. As shown, the distribution is extremely uneven. The samples of the top 10 most frequent relationships account for almost 90\% samples, while those of the rest 40 relationships merely account for about 10\%. Thus, the R@$K$ metric is dominated by the performance of these most frequent relationships. As shown in Figure \ref{fig:sg-comparison}, current state-of-the-art method (i.e., SMN) performs quite well for these relationships such as ``on'', ``has''; thus it can achieve a good R@$K$. However, SMN performs quite poorly for the relationships that have fewer samples (e.g., ``make of'', ``to''). The mR@$K$ metric measures the overall performance over all relationships; thus these poor results lead to an obvious drop on this metric. Different from existing methods, our model integrates prior knowledge to explicitly regularize the semantic space; thus it also performs well for these less frequent relationships. In this way, our model can well address the issue of uneven distribution of relationships.

To present a more direct comparison of the relation between the performance improvement and sample number, we further present the R@50 improvement for each relationship and sample proportion in Figure \ref{fig:sg-improvement} and \ref{fig:sg-scatter}. As shown, our model achieves evident improvement in almost all relationships (47/50). Besides, the improvement is more obvious for the relationships with fewer samples.

\subsection{Ablative study}
The core of our method is the explicit incorporation of statistical correlation of object pair and their relationship. To better verify its effectiveness, we replace the statistical probabilities with uniform distribution, i.e., assigning each $m_{cc'k}$ to $\frac{1}{K}$, leaving other components unchanged. The experiment is conducted on the VG dataset and the results are presented in Table \ref{table:vg-mrk-ablation}. We find that the mean mR decreases from 11.7\% to 7.9\% and the mean R decreases from 44.1\% to 40.6\%. This obvious performance drop clearly indicates incorporating statistical correlations significantly helps scene graph generation. 

It is another important module that our method propagates messages through regions appearing in the image to learn contextualized representation. Similarly, we analyze its significance by replacing the statistical probabilities with a uniform distribution, and retrain the model on the VG dataset. As shown in Table \ref{table:vg-mrk-ablation}, both the mean mR and mean R suffer from 0.3\% drop.

\section{Conclusion}
The prior knowledge of statistical correlations between object pair and their relationship can help regularize the semantic space of relationship prediction given target object pair, and thus effectively address the issue of the uneven distribution over different relationships. In this work, we show these correlations can be explicitly represented by a knowledge graph, in which a routing mechanism is learned to propagate node messages through the graph under the explicit guidance of the structured knowledge. We conduct experiments on the most widely used Visual Genome benchmark and demonstrate the superiority of the proposed method.

{\small
\bibliographystyle{ieee}
\bibliography{egbib}
}

\end{document}